\documentclass[runningheads]{llncs}
\usepackage[colorlinks=true,linkcolor=red,citecolor=green]{hyperref}

\usepackage[width=122mm,left=12mm,paperwidth=146mm,height=193mm,top=12mm,paperheight=217mm]{geometry}

\usepackage{graphicx}
\usepackage{tikz}
\usepackage{comment}
\usepackage{amsmath,amssymb} 
\usepackage{color}
\usepackage[accsupp]{axessibility} 
\usepackage{graphicx}
\usepackage{amsmath}
\usepackage{amssymb}
\usepackage{booktabs}
\usepackage{multicol}
\usepackage{multirow}
\usepackage{subfigure}
\usepackage{colortbl}
\usepackage{enumitem}
\usepackage[normalem]{ulem}
\usepackage[T1]{fontenc}
\usepackage{dsfont}
\usepackage{ragged2e}
\usepackage{cite}
\usepackage{caption}
\usepackage{mathrsfs}
\usepackage{mathtools}
\usepackage{algorithm2e}
\usepackage{algpseudocode}
\usepackage{tabularx}
\usepackage{caption}
\usepackage{ragged2e}
\usepackage{epsfig}
\usepackage{tensor}
\usepackage{booktabs}
\usepackage{balance}
\usepackage{wrapfig}
\usepackage{lipsum}
\usepackage{listings}
\usepackage{xcolor}
\usepackage{subfiles}
\pdfoutput=1
\definecolor{Gray}{gray}{0.9}
\usepackage[capitalize]{cleveref}

\crefname{section}{Sec.}{Secs.}
\Crefname{section}{Section}{Sections}
\Crefname{table}{Table}{Tables}
\crefname{table}{Tab.}{Tabs.}
\newcommand{\nrsfm}{{{NRS\emph{f}M}}}
\newcommand{\etal}{\textit{et al.}}
\newcommand{\ie}{\textit{i.e.}}
\newcommand{\eg}{\textit{e.g.}}
\newcommand{\formattedparagraph}[1]{\noindent \textbf{#1}}

\definecolor{navyblue}{rgb}{0.85,0.95,1.0}
\RestyleAlgo{ruled}
\definecolor{amber}{rgb}{1.0, 0.49, 0.0}
\definecolor{ao}{rgb}{0.0, 0.5, 0.0}
\definecolor{darklavender}{rgb}{0.45, 0.31, 0.59}
\definecolor{lavender}{rgb}{0.71, 0.49, 0.86}
\definecolor{sangria}{rgb}{0.57, 0.0, 0.04}

\begin{document}

\pagestyle{headings}
\mainmatter
\title{Organic Priors in Non-Rigid Structure from Motion}
\titlerunning{Organic Priors in Non-Rigid Structure from Motion} 
\authorrunning{Suryansh Kumar and Luc Van Gool} 
\author{Suryansh Kumar$^1$ \quad \quad \quad \quad \quad Luc Van Gool$^{1, 2}$\\
{\tt\small \quad \quad sukumar@vision.ee.ethz.ch \quad vangool@vision.ee.ethz.ch
}
}
\institute{$^{1}$ETH Z\"urich Switzerland, $^2$KU Leuven Belgium}

\maketitle

\begin{abstract}
\justify
This paper advocates the use of organic priors in classical non-rigid structure from motion (\nrsfm). By organic priors, we mean invaluable intermediate prior information intrinsic to the \nrsfm ~matrix factorization theory. It is shown that such priors reside in the factorized matrices, and quite surprisingly, existing methods generally disregard them. The paper's main contribution is to put forward a simple, methodical, and practical method that can effectively exploit such organic priors to solve \nrsfm. The proposed method does not make assumptions other than the popular one on the low-rank shape and offers a reliable solution to \nrsfm ~under orthographic projection. Our work reveals that the accessibility of organic priors is independent of the camera motion and shape deformation type. Besides that, the paper provides insights into the \nrsfm ~factorization---both in terms of shape and motion---and is the first approach to show the benefit of single rotation averaging for \nrsfm. Furthermore, we outline how to effectively recover motion and non-rigid 3D shape using the proposed organic prior based approach and demonstrate results that outperform prior-free \nrsfm ~performance by a significant margin. Finally, we present the benefits of our method via extensive experiments and evaluations on several benchmark datasets. 
\end{abstract}

\keywords Organic Priors, Non-Rigid Structure from Motion, Rank Minimization, Rotation Averaging, Matrix Factorization.

\section{Introduction}
Non-rigid structure from motion (\nrsfm) factorization is a classical problem in geometric computer vision \cite{bregler2000recovering,1885-164278}. The problem's primary objective is to recover 3D shape of a deforming object from a given set of image key-points tracked across multiple images. As a result, it is sometimes referred as solving an inverse graphics problem \cite{salzmann2010deformable}. An effective solution to \nrsfm ~is of significant importance to many computer vision and geometry processing applications\cite{matthews20072d,bronstein2008numerical}.

It is now widely accepted that the \nrsfm ~problem is challenging to work out if the shape deforms arbitrarily across images, as it becomes equivalent to a non-rigid shape recovery problem using a single image at a time, which is ill-posed. Accordingly, several assumptions and priors are often used to make the problem solvable and computationally tractable. For instance, the deforming shape spans a low-rank space\cite{dai2014simple}, smooth temporal shape deformation \cite{bartoli2008coarse,aanaes2002estimation}, shape or trajectory lies in the union of linear subspace \cite{zhu2014complex, kumar2016multi, kumar2017spatio, kumar2019jumping, kumar2018scalable, kumar2020dense} and the local surface deformation is rigid or near rigid \cite{taylor2010non,lee2016consensus,fayad2010piecewise,russell2011energy}. Other favored prior assumptions include smooth camera motion \cite{rabaud2008re,kumar2020non}, a piece-wise planar deformation model \cite{fayad2010piecewise, kumar2019superpixel, kumar2017monocular, kumar2019dense}, a Gaussian shape prior distribution \cite{torresani2004learning}, the availability of a 3D shape template \cite{salzmann2010deformable}, and shapes across frames must align \cite{lee2013procrustean}. Despite that, \nrsfm ~remains a challenging and active research problem.

Meanwhile, there exist several popular methods to solve \nrsfm ~\cite{bregler2000recovering,torresani2004learning,akhter2009nonrigid,dai2014simple,lee2013procrustean}. Here, we will concern ourselves with the theory of matrix factorization for \nrsfm ~elaborated in 1999-2000 by Bregler \etal  \cite{bregler2000recovering}\footnote{See, however, \textit{C. Tomasi and T. Kanade}, \textbf{pp. 137-154}, IJCV (1992) for the original matrix factorization theory for shape and motion estimation, although devoted to the rigid S\textit{f}M problem \cite{tomasi1992shape}.}. It is a simple yet an effective approach to solve \nrsfm. In the context of matrix factorization, one of the commonly used prior assumptions is that the non-rigid shape spans a low-rank space \ie, the shape at each instance can be represented as a linear combination of a small set of basis shapes \cite{bregler2000recovering}. This paper adheres to such an assumption and shows that other important prior information resides within the \emph{intermediate} factorized matrices ---termed as organic priors. Surprisingly, most existing methods, if not all, ignore them. We used the word \textcolor{red}{{{\textless\textless}}}\texttt{organic}\textcolor{red}{{{\textgreater\textgreater}}} because they come naturally by properly conceiving the algebraic and geometric construction of \nrsfm ~factorization \cite{tomasi1992shape,bregler2000recovering,dai2014simple}. Furthermore, this paper contends that the use of external priors and assumptions not only restricts the practical use of \nrsfm ~methods, but also constrains the understanding and broader use of the well-known theory \cite{dai2014simple}. Yet, unlike \cite{dai2014simple}, we advocate the use of organic priors, which is predicated on the proposition put forward by Kumar \cite{kumar2020non}. In this paper, we will show how to procure organic priors and exploit them effectively.

One of the critical innovations in \nrsfm ~factorization that disputed the use of extraneous priors was introduced in \cite{dai2014simple, dai2012simple}. The algorithm proposed in that paper does not use any prior other than the low-rank shape assumption. Nevertheless, despite its theoretical elegance and challenging argument, it fails to perform well on benchmark datasets \cite{jensen2018benchmark,akhter2009nonrigid,torresani2008nonrigid,garg2013dense}. Recently, Kumar \cite{kumar2020non} highlighted the possible reasons and exploited its missing pieces to gain performance. It was shown that a better rotation and shape could be estimated using the prior-free method's theory \cite{dai2012simple, dai2014simple}. Still, \cite{kumar2020non} based his work on a smooth camera motion assumption that requires a brute force, heuristic search in the rotation space. In contrast, this paper puts forward a systematic method for \nrsfm ~factorization that encourages the use of organic priors extracted from the factorized matrices. Experiments on synthetic and real benchmarks show that our approach consistently provides excellent 3D reconstruction results. This indicates the strength of matrix factorization theory for \nrsfm. In summary, our contributions are

\begin{itemize}[leftmargin=*,topsep=0pt, noitemsep]
    \item [\textcolor{black}{$\bullet$}] A methodical approach for solving \nrsfm ~that provides outstanding results using simple matrix factorization idea under the low-rank shape assumption.
    \item [\textcolor{black}{$\bullet$}] An approach that endorses the use of organic priors rather than extraneous priors or assumptions. Our method introduce single rotation averaging to estimate better rotation while being free from smooth camera motion heuristics.
    \item [\textcolor{black}{$\bullet$}] A different setup for low-rank shape optimization is proposed. We present a blend of partial sum minimization of singular values theory and weighted nuclear norm optimization for shape recovery. We observed that the proposed optimization better exploits the organic shape priors and yields shape reconstructions superior to other popular \nrsfm ~factorization methods \cite{iglesias2020accurate,kumar2020non,dai2014simple}.
\end{itemize}

\noindent
Further, we proffer the benefits of $L_1$ single rotation averaging for \nrsfm ~factorization, which is excellent at providing robust rotation solutions. Although most of \nrsfm ~factorization focuses on sparse key-point sets, our method is equally effective for dense feature points and compares favorably with well crafted state-of-the-art dense \nrsfm ~methods such as \cite{garg2013dense,kumar2018scalable,kumar2019jumping}.

\section{Overview and Key Strategy}
\formattedparagraph{General Definition and Classical Setup.} In \nrsfm, a measurement matrix $\mathbf{W} \in \mathbb{R}^{2F \times P}$ is defined as a matrix containing the image coordinates ($\mathbf{w}_{f, p} \in \mathbb{R}^{2 \times 1}$) of $P$ feature points tracked across $F$ image frames. $\mathbf{W}$ is generally mean-centered and given as an input to the factorization method \cite{bregler2000recovering}. Under an orthographic camera model assumption, the \nrsfm ~factorization theory proposes to decompose the $\mathbf{W}$ into a product of a rotation matrix $\mathbf{R} \in \mathbb{R}^{2F \times 3F}$ and a non-rigid shape matrix $\mathbf{X} \in \mathbb{R}^{3F \times P}$ such that $\mathbf{W} \approx \mathbf{RX}$.

A practical method for \nrsfm ~factorization was initially proposed by Bregler \etal \cite{bregler2000recovering}. Using the linear model proposition, a non-rigid shape $\mathbf{X}_i \in \mathbb{R}^{3 \times P}$ for $i^{th}$ frame was represented as a linear combination of $K$ basis shapes $\mathbf{B}_{k} \in \mathbb{R}^{3 \times P}$ \ie,  $\mathbf{X}_i = \sum_{k=1}^{K}c_{ik}\mathbf{B}_{k}$, where $c_{ik}$ denotes the shape coefficients. Using such a shape representation, the $\mathbf{W}$ matrix is decomposed as follows:

\begin{equation}\label{eq:2}
\small
\begin{aligned}
& \displaystyle  ~~~~~\mathbf{W} =\begin{bmatrix}
    \mathbf{w}_{11} \dots \mathbf{w}_{1P}\\
    \dots\\
   \mathbf{w}_{F1} \dots \mathbf{w}_{FP}
\end{bmatrix}=\begin{bmatrix} \mathbf{R}_1 \mathbf{X}_1 \\..\\ \mathbf{R}_{F}  \mathbf{X}_{F} \end{bmatrix} =\begin{bmatrix}
    \mathbf{c}_{11} \mathbf{R}_1 \dots \mathbf{c}_{1 K} \mathbf{R}_1\\
    \dots\\
   \mathbf{c}_{F 1} \mathbf{R}_{F} \dots \mathbf{c}_{F K} \mathbf{R}_{F}
\end{bmatrix}\begin{bmatrix} \mathbf{B}_1 \\ .. \\  \mathbf{B}_{ K} \end{bmatrix} \\
& \displaystyle \Rightarrow \mathbf{W} =  \mathbf{R}( \mathbf{C} \otimes  \mathbf{I}_{3})  \mathbf{B} = \mathbf{M}\mathbf{B}
\end{aligned}
\end{equation}
where, $\mathbf{R}_{i} \in \mathbb{R}^{2 \times 3}$ denotes the $i^{th}$ frame rotation matrix, and $\otimes$ the Kronecker product. $\mathbf{M} \in \mathbb{R}^{2F \times 3K}$, $\mathbf{B} \in \mathbb{R}^{3K \times P}$ and $\mathbf{I}_3$ is the $3 \times 3$ identity matrix. It is easy to infer from the above construction that $\texttt{rank}(\mathbf{W}) \leq 3K$. 

Since there is no general way to solve for $\mathbf{R}, \mathbf{C}$, and $\mathbf{B}$ directly, rank $3K$ factorization of $\mathbf{W}$ via its Singular Value Decomposition (\texttt{svd}) gives a natural way to solve the problem under the orthonormality constraint of the rotation space \cite{bregler2000recovering,akhter2009nonrigid,dai2014simple}. As it is well-known that factorization of $\mathbf{W}$ via \texttt{svd} is not unique \cite{tomasi1992shape,bregler2000recovering}, there must exist a corrective matrix $\mathbf{G} \in \mathbb{R}^{3K \times 3K}$ such that $\mathbf{W} = (\mathbf{\hat{M}} \mathbf{G}) (\mathbf{G}^{-1} \mathbf{\hat{B}}) = \mathbf{M} \mathbf{B}$. And therefore, once the \texttt{svd} of $\mathbf{W}$ is computed, a general rule of thumb in \nrsfm ~factorization is to first solve for the $\mathbf{G}$ matrix, followed by the estimation of $\mathbf{R}$ and $\mathbf{X}$, respectively \cite{akhter2009nonrigid,dai2014simple}.

\begin{figure*}[t]
\centering
\includegraphics[width=1.0\textwidth] {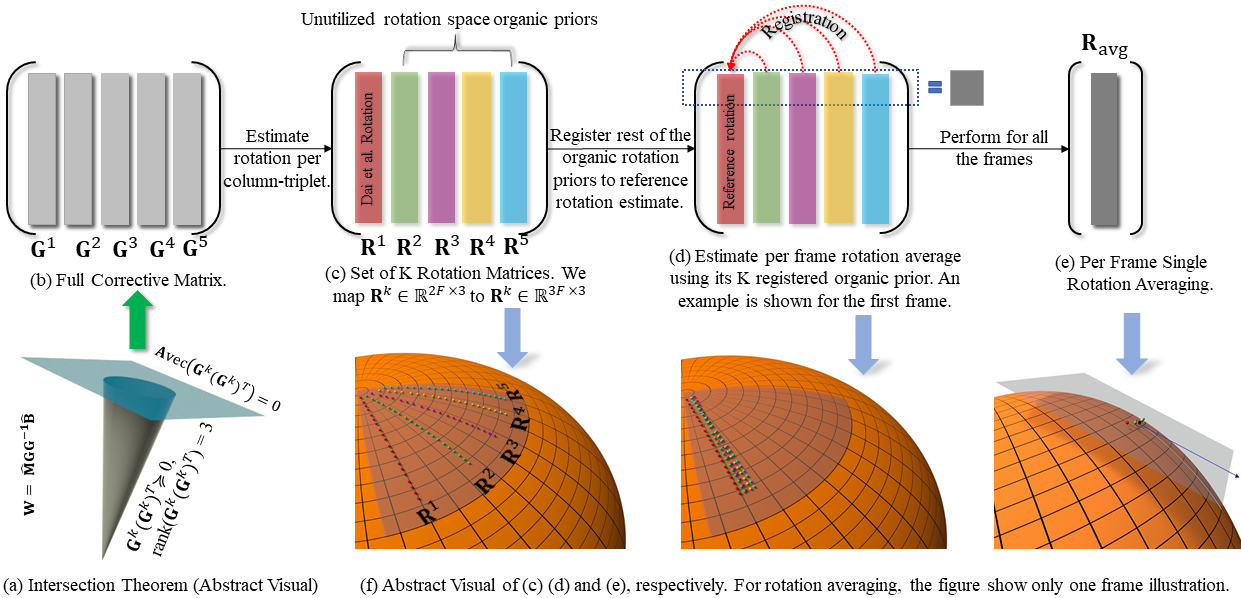}
\caption{A visual illustration of our rotation estimation approach for $K=5$. (a)-(b) Use the \cite{dai2014simple} intersection theorem to recover all $\mathbf{G}^{k}$s'. (b)-(c) Recover $\mathbf{R}^{k} \in \mathbb{R}^{2F \times 3}$ corresponding to each column triplet. (c) Map per frame $2 \times 3$ rotation to ${SO}(3)$ via a cross product, taking care of the determinant sign. (d) Register $(K-1)$ rotation to the reference rotation (e) Perform per frame single rotation averaging to recover $\mathbf{R}_\text{avg}$.}
\label{fig:rotation_concept}
\end{figure*}

\smallskip
\formattedparagraph{(a) Background on Corrective and Rotation Matrix Estimation.} To solve for $\mathbf{G}$, orthonormality constraints are enforced \cite{bregler2000recovering,akhter2009defense}. Few works proposed in the past solve for the full $\mathbf{G} \in \mathbb{R}^{3K \times 3K}$ matrix (\ie, for all its matrix entries) to estimate the rotation matrix \cite{brand2005direct,xiao2004closed,akhter2009defense,valmadre2015closed}. In contrast, Dai \etal \cite{dai2014simple} argued that rather than solving for the full $\mathbf{G}$, simply solve for $\mathbf{G}^{1} \in \mathbb{R}^{3K \times 3}$ (first 3 columns or first corrective triplet) leveraging Akhter \etal's \cite{akhter2009defense} theory. Yet, there exist $K$ such triplets (see Fig.\ref{fig:rotation_concept}). Even if we don't deviate from \cite{akhter2009defense} theory, the question that still remains with the use of  \cite{dai2014simple} rotation estimation theory is:

\smallskip
\noindent
\emph{Q1. {Do we utilize all possible rotations that can be recovered from  \cite{dai2014simple} rotation estimation theory?}} The answer is no, as recently argued by Kumar \cite{kumar2020non}. He proposed to inspect all $K$ column triplets in the corrective matrix ($\mathbf{G}$), and recover $K$ possible rotation solutions $\mathbf{R}^{k} \in \mathbb{R}^{2F \times 3}$, where $k \in \{1, 2,\dots, K\}$. He then selected the one $\mathbf{R}^{k}$ that provides a smooth camera motion trajectory. Yet, this solution is heuristic in nature and requires a qualitative inspection of all the $K$ rotations. The point to note is that, similar to Dai \etal \cite{dai2014simple}, Kumar's \cite{kumar2020non} solution at the end does not fully utilize all the $K$ rotation priors and eventually ends up aborting the rest of the near smooth or non-smooth $(K-1)$ solutions. We call those $(K-1)$ rotation solutions 
\textcolor{red}{{{\textless\textless}}}\texttt{organic priors in the rotation space}\textcolor{red}{{{\textgreater\textgreater}}}.
Our proposed method utilizes all those organic rotation priors to estimate a better and more informed rotation matrix.

\smallskip
\formattedparagraph{(b) Background on the Shape Matrix Estimation.} After solving for the rotation, the goal is to estimate the shape with the rank $K$ constraint. Generally, an initial solution to the shape can be estimated in a closed form using $\mathbf{X}_{init} = {\texttt{pinv}} (\mathbf{R})\mathbf{W}$\footnote{\texttt{pinv()} symbolizes Moore–Penrose inverse of a matrix, also known as pseudoinverse.}. Yet, this may produce a planar solution as outlined in \cite{valmadre2015closed}. In spite of that, $\mathbf{X}_{init}$ provides useful information about the true shape and can be used as a shape variable ($\mathbf{X}$) initialization in the following rank-optimization problem:
\begin{equation}
    \begin{aligned}
    & \displaystyle \underset{\mathbf{X}^{\sharp}} {{\textrm{minimize}}} \frac{1}{2}\sum_{i=1}^{F} \sum_{j=1}^{P}\|\mathbf{w}_{ij} - \mathbf{R}_{i}\mathbf{x}_{ij}\|^{2},
    ~{\textrm{s.t.}} ~\textrm{rank}(\mathbf{X}^{\sharp}) \leq K
    \end{aligned}\label{eq:shape_initial_form}
\end{equation}
where, $\mathbf{x}_{ij}$ denote the 3D point $j$ in the $i^{th}$ view, and $\mathbf{w}_{ij}$ is its corresponding projection. $\mathbf{X}^{\sharp} \in \mathbb{R}^{F \times 3P}$ is the reshape of the shape matrix ($\mathbf{X}$) for $K$ shape basis constraint \cite{akhter2011trajectory,dai2014simple}. 

There exist several approaches to solve the Eq.\eqref{eq:shape_initial_form} optimization \cite{akhter2011trajectory,gotardo2011non,paladini2009factorization,dai2014simple}. Among them, relaxed rank-minimization via appropriate matrix-norm minimization is widely used to recover a low-rank shape matrix providing favorable accuracy and robust results \cite{dai2014simple,garg2013dense,valmadre2015closed,kumar2020non,iglesias2020accurate}. In this paper, we exploit the singular values of the $\mathbf{X}_{init}$, which we call \textcolor{red}{{{\textless\textless}}}\texttt{organic priors in the shape space}\textcolor{red}{{{\textgreater\textgreater}}} to recover better solution than the recent state-of-the-art \cite{iglesias2020accurate,ornhag2021bilinear}. Although Kumar \cite{kumar2020non} work is the first to propose and utilize such priors for better shape reconstructions, in this paper we show that we can do better\footnote{Familiarity with \cite{dai2014simple, kumar2020non} gives a good insight on our paper's novelty.}. This brings us to the next question:

\smallskip
\noindent
\emph{Q2. {Can we make better use of the organic shape prior to solve for the shape matrix?}} We will show that we can. When solving the relaxed rank minimization optimization problem of the shape matrix \cite{dai2014simple}, it is not beneficial to equally penalize all the singular values of the shape matrix. Hence, for effective shape recovery, one can use the $\mathbf{X}_{init}$ singular values prior to regularize the shape matrix rank-optimization \cite{kumar2020non}. In particular, perform a weighted nuclear norm (WNN) minimization of the shape matrix, and assign the weights to the shape variable that is  inversely proportional to the $\mathbf{X}_{init}$ singular values magnitude \cite{kumar2020non}. In this paper, we go a step further. We propose to preserve the first component of the shape during its WNN minimization, \ie, ~to not penalize the first singular value shape prior from $\mathbf{X}_{init}$.  We empirically observed that the first singular value of $\mathbf{X}_{init}$, more often than not, does contain rich information about the true shape. Penalizing the first singular value during WNN minimization of the shape matrix may needlessly hurts the overall results. Consequently, we introduce a mix of partial sum minimization of singular values \cite{oh2016partial} and WNN minimization \cite{kumar2020non} to recover a better shape matrix estimate.

\section{Proposed Approach}
First, we provide details of our approach to solve for the rotation matrix, followed by the shape matrix estimation. 

\subsection{Rotation Estimation}\label{ss:rotation_estimation}
\noindent
To put our work in context, we highlight some previous efforts that took a similar direction towards enhancing the rotation estimate for \nrsfm.

\smallskip
\formattedparagraph{Relation to previous methods.} As mentioned before, there exist $K$ corrective column triplets in the $\mathbf{G}$ matrix (Fig.\ref{fig:rotation_concept}(b)). Brand \cite{brand2005direct} and Akhter \etal \cite{akhter2009defense} solves for all corrective triplets jointly. Xiao \etal \cite{xiao2004closed} proposed to independently solve for each corrective triplet ($\mathbf{G}^{k}$) and align $\mathbf{G}^{k}$'s using the Procrustes method up to sign\cite{rabaud2008re}. Lee \etal \cite{lee2013procrustean} proposed an additional constraint on the rotation by posing \nrsfm ~as a shape alignment problem. By comparison, the Dai \etal \cite{dai2014simple} method is a simple and effective way to compute rotation. It estimates only the $1^{st}$ column-triplet of $\mathbf{G}$ \ie, $\mathbf{G}^{1}$ to recover $\mathbf{R}$ (see Fig.\ref{fig:rotation_concept}).


On the contrary, we propose to first compute all the column-triplets \ie, $\mathbf{G}^{k}$~$\forall ~k \in \{1, 2,\dots, K\}$ and their corresponding rotation matrices $\mathbf{R}^{k} \in \mathbb{R}^{2F \times 3}$, using Dai \etal \cite{dai2014simple}. Later, we use all $K$ rotation estimates per frame to estimate a better rotation matrix $\mathbf{R} \in \mathbb{R}^{2F \times 3F}$ via the theory of single rotation averaging \cite{hartley2013rotation}. Rather than aligning $\mathbf{G}^{k}$s as in \cite{akhter2009defense,xiao2004closed}, we register the rotations. Our approach consists of the following steps applied in sequel:

\begin{itemize}[leftmargin=*,topsep=0pt, noitemsep]
    \item [\textcolor{amber}{$\bullet$}] Recover $\mathbf{G}^{k}$ and its corresponding $\mathbf{R}^{k} \in \mathbb{R}^{2F \times 3}$ using \cite{dai2014simple},  $\forall ~k \in \{1,\dots, K\}$.
    \item [\textcolor{ao}{$\bullet$}] Map $\mathbf{R}^{k} \in \mathbb{R}^{2F \times 3}$ $\mapsto$ $\mathbf{R}^{k} \in \mathbb{R}^{3F \times 3}$ via the cross product of per frame $2 \times 3$ orthographic rotation estimates, while correcting for the sign of the determinant, if applicable.
    \item [\textcolor{darklavender}{$\bullet$}] Take the rotation due to the first column-triplet \ie, $\mathbf{R}^{1} \in \mathbb{R}^{3F \times 3}$ as the reference rotation matrix and register the other $(K-1)$ \ie, $\mathbf{R}^{2}$ to $\mathbf{R}^{K}$ rotation estimates to it. After registration, filter the rotation sample if the distance w.r.t its reference rotation sample is greater than $\delta$ \cite{hartley2013rotation} (see below for details).
    \item [\textcolor{sangria}{$\bullet$}] Perform per frame single rotation averaging of all the aligned rotation priors to recover $\mathbf{R}_{avg} \in \mathbb{R}^{3F \times 3}$. Later, convert $\mathbf{R}_{avg}$ per frame to orthographic form and place it in the block diagonal structure to construct $\mathbf{R} \in \mathbb{R}^{2F \times 3F}$.
\end{itemize}

\noindent
Before performing single rotation averaging, we align all rotation priors due to the global ambiguity (see Fig.\ref{fig:rotation_concept}\textcolor{red}{(b)}-Fig.\ref{fig:rotation_concept}\textcolor{red}{(c)} visual). We align the other $(K-1)$ rotations to $\mathbf{R}^{1}$ using the following optimization.
\begin{equation}
    \begin{aligned}
     & \displaystyle 
     \underset{R_\text{reg}^{k}} {{\textrm{minimize}}} \sum_{f=1}^{F}\|\mathbf{R}_f^{1} - \mathbf{R}_f^{k}({\mathbf{R}_\text{reg}^{k}})^{{T}}\|_\mathcal{F}^2; ~{\textrm{s.t.}} ~\mathbf{R}_\text{reg}^{k} \in {SO}(3), ~\forall ~k \in \{2,\dots,K\}
    \end{aligned}\label{eq:motion_registration}
\end{equation}
where, $k\in \mathbb{Z}$. In the paper, $\|.\|_\mathcal{F}$ denotes the Frobenius norm. Using Eq.\eqref{eq:motion_registration} optimization, we recover $(K-1)$ $\mathbf{R}_\text{reg}^{k} \in \mathbb{R}^{3 \times 3}$ to register the organic rotation priors for averaging. Next, we perform single rotation averaging per frame.

\smallskip
\formattedparagraph{Single rotation averaging.} Given a set of $n \geq 1$ rotations $\{R_1, R_2,\dots, R_n\} \subset {SO}(3)$, the goal of single rotation averaging (SRA) is to find the average of a set of rotations \cite{hartley2013rotation}. It can also be conceived as finding a rotation sample on the ${SO}(3)$ manifold that minimizes the following cost function
\begin{equation}\label{eq:single_ravg}
     \underset{R \in {SO}(3)}{\text{argmin}} \sum_{i=1}^{n} d^{p}(R_{i}, R)
\end{equation}
$d()$ denotes a suitable metric function. We use $p=1$ for its robustness and accuracy as compared to $p=2$ \cite{hartley2011l1}. For our problem, we have $K$ rotation samples for each frame (see Fig.\ref{fig:rotation_concept}\textcolor{red}{(c)}). Accordingly, we modify Eq.\eqref{eq:single_ravg} as:
\begin{equation}\label{eq:single_ravg_modi}
     \underset{\mathbf{R}_f \in {SO}(3)}{\text{argmin}} \sum_{k=1}^{K} d^1(\Tilde{\mathbf{R}}_{f}^{k}, \mathbf{R}_{f})
\end{equation}
Here, $\Tilde{\mathbf{R}}_{f}^{k}$ is the $k^{th}$ registered rotation for the $f^{th}$ frame, \ie, averaging across rows after registration (Fig.\ref{fig:rotation_concept}\textcolor{red}{(c)}). We solve Eq.\eqref{eq:single_ravg_modi} for all the frames using the Weiszfeld algorithm for $L_1$ rotation averaging \cite{hartley2011l1}. The average is computed in the local tangent space of ${SO}(3)$ centered at the current estimate and then back-projected onto ${SO}(3)$ using the exponential map (Fig.\ref{fig:rotation_concept}\textcolor{red}{(d)}-Fig.\ref{fig:rotation_concept}\textcolor{red}{(e)}). Yet, instead of initializing using the chordal $L_2$ mean, we use the starting point using the following equation proposed recently by Lee and Civera \cite{lee2020robust}.
\begin{equation}\label{eq:robust_initialization}
    \mathbf{S}_o = \underset{\mathbf{S} \in \mathbb{R}^{3 \times 3}}{\text{argmin}} \sum_{i=1}^{K} \sum_{j=1}^{3} \sum_{k=1}^3 \big|(\mathbf{R}_i - \mathbf{S})_{jk}\big|_1
\end{equation}
Eq.\eqref{eq:robust_initialization} is an element-wise $L_1$ norm matrix entries, minimizing the sum of absolute differences from the $\mathbf{R}_{i}$ at $(j, k)$ value \ie, $(\mathbf{S_o})_{jk} = \textrm{median}(\{\mathbf{R}_{i}\}_{i=1}^{K}) ~\forall ~j, k \in \{1, 2, 3\}$. After median computation, its equivalent rotation representation is obtained by projecting $\mathbf{S}_{o}$ onto ${SO}(3)$ using $\Upsilon$ operator. For $\Psi \in \mathbb{R}^{3 \times 3}$ matrix, we define $\Upsilon_{{SO}(3)}(\Psi) = \mathbf{U}\mathbf{D}\mathbf{V}^T$, where $\mathbf{U}\mathbf{D}\mathbf{V}^T$ is \texttt{svd} of $\Psi$ and $\mathbf{D} = \texttt{diag}(1, 1, -1)$ if $\det(\mathbf{U}\mathbf{V}^{T}) < 0$ or $\mathbf{I}_{3 \times 3}$ otherwise.

\setlength{\intextsep}{5pt}%
\setlength{\columnsep}{5pt}%
\begin{wrapfigure}[14]{R}{0.55\textwidth}
\scriptsize
      \begin{algorithm}[H]                
        \SetCustomAlgoRuledWidth{0.45\textwidth}  
       \caption{\scriptsize $L_1$ Single Rotation Avg.}\label{alg:single_motion_avg_algo}
          \KwData{Set of rotation $\{\textbf{R}_{i}\}_{i=1}^{K}$, ~$\epsilon_t = 1e^{-3}$}
          \KwResult{$L_1$ mean \ie, median rotation}
          Set $\mathbf{S}_o := \text{median}(\{\mathbf{R}_{i}\}_{i=1}^{K})$;\\
          \textcolor{sangria}{/*Project median on ${SO}(3)$*/}\\
          Set $\mathbf{R}_o := \Upsilon_{{SO}(3)}(\mathbf{S}_o)$; 
          Set $\mathbf{R}_\text{avg} := \mathbf{R}_o$;\\
           \While{}{ $\mathbf{v}_{i} := \log(\mathbf{R}_{i}\mathbf{R}_\text{avg}^{T}) ~\forall ~i = {1, 2.., K}$;\\
  $\Delta \mathbf{v} := \frac{\sum_{i=1}^{K}\mathbf{v}_i/\|\mathbf{v}_i\|}{\sum_{i=1}^{K} 1/\|\mathbf{v}_i\|}$;\textcolor{sangria}{/*Weiszfeld step*/}\\
  $\mathbf{R}_\text{avg} := \exp(\Delta \mathbf{v}) \mathbf{R}_\text{avg}$\\
  \If{$\| \Delta \mathbf{v}\| < \epsilon_t$}{
  \textbf{break};
  }
 }
\textbf{return} $\mathbf{R}_\text{avg}$;\\
      \end{algorithm}
\end{wrapfigure}

\textbf{Algorithm} \ref{alg:single_motion_avg_algo} provide our implementation for single rotation averaging. Empirically, after registration and minor filtering of the rotation samples\footnote{filter if sample is too far to the reference rotation after registration.}, we observed per frame rotation samples are reasonably close, which is good for convergence \cite{hartley2013rotation}.  Averaging per frame rotation priors, we recover $\mathbf{R}_\text{avg} \in \mathbb{R}^{3F \times 3}$. For more details on single rotation averaging and its convergence analysis refer \cite{hartley2013rotation,lee2020robust}.

To compute $\mathbf{R} \in \mathbb{R}^{2F \times 3F}$ from $\mathbf{R}_\text{avg} \in \mathbb{R}^{3F \times 3}$, we take $\mathbf{R}_\text{avg}$'s per frame $3 \times 3$ matrix, drop its  $3^\text{rd}$ row  and place it to the diagonal of $\mathbf{R}$, and perform this step for all frame\footnote{After registration, if samples are filtered out due to its distance from the reference rotation (more than $\delta$), then per frame rotations is less than $K$.}.

\subsection{Shape Estimation}\label{ss:shape_estimation}
Once we estimated the rotation matrix, our goal is to recover the shape matrix. An easy way to compute shape is $\mathbf{X}_{init} = \texttt{pinv}(\mathbf{R})\mathbf{W}$, which is consistent with the assumption of low rank shape matrix and it minimizes the re-projection error. To show the merit of our rotation estimation, we tabulate the pseudo inverse shape reconstruction result using our rotation compared to BMM \cite{dai2014simple} in Tab.(\ref{tab:pinv_soln}). Clearly, our rotation improves the $\mathbf{X}_{init}$, \ie, the initial shape solution, by a large margin. However, $\mathbf{X}_{init}$ may be a 3D reconstruction co-planar in every frame and ``the catch is that there are in fact many solutions which minimize the rank''  \cite{valmadre2015closed}. Therefore, further optimization of the shape matrix is recommended. Let's review Dai \etal's \cite{dai2014simple} relaxed rank-minimization and recent improvement over it to  better place our approach.
%

\begin{table}[t]
    \scriptsize
    \centering
    \resizebox{\columnwidth}{!}
    {
    \begin{tabular}{c|c|c|c|c|c|c|c|c}
    \hline
    \rowcolor[gray]{0.75}
         Dataset & Drink & Pickup & Yoga & Stretch & Dance & Face & Walking & Shark\\
         \hline
         BMM-PI \cite{dai2014simple} & 0.4449 & 0.5989 & 0.6523 & 0.4784 & 0.5764 & 0.4848 & 0.5100 & 0.8784\\
         \hline
    \rowcolor{red!15}
         Ours-PI & \textbf{0.2195} & \textbf{0.2985} & \textbf{0.2740} & \textbf{0.2238} & \textbf{0.3014} & \textbf{0.2995} & \textbf{0.2702} & \textbf{0.3053} \\
         \hline
    \end{tabular}
    }
    \caption{Pseudo inverse (PI) shape results comparison with BMM \cite{dai2014simple} via $e_{3d}$ metric. Compared to \cite{dai2014simple}, our approach dramatically improves PI shape accuracy, showing the benefit of using rotation organic prior. $e_{3d}$ definition is provided in Sec.\S \ref{sec:experiments} }
    \label{tab:pinv_soln}
\end{table}

\formattedparagraph{Relation to previous methods.} Given rotation matrix $\mathbf{R} \in \mathbb{R}^{2F \times 3F}$, Dai \etal \cite{dai2014simple} perform the following optimization for low-rank shape matrix estimation.
\begin{equation}\label{eq:dai_opt}
\begin{aligned}
& \displaystyle \underset{ \mathbf{X}^{\sharp},  \mathbf{X}} {{\textrm{minimize}}} ~\frac{1}{2}\|\mathbf{W} - \mathbf{R}\mathbf{X}\|_\mathcal{F}^{2} + \mu\| \mathbf{X}^{\sharp}\|_*; {\textrm{subject to:}}~ \mathbf{X}^{\sharp} = \mathbf{\Phi}(\textbf{X})
\end{aligned}
\end{equation}
Here, $\mathbf{\Phi}(.)$ is a function that maps $\mathbf{X} \in \mathbb{R}^{3F \times P}$ to $\mathbf{X}^{\sharp} \in \mathbb{R}^{F \times 3P}$. $\mu$ is a scalar constant and $\|. \|_*$ denotes the nuclear norm of the matrix, which is a convex bound of the matrix rank and can give a good solution to rank minimization problems under restricted isometry property constraints \cite{recht2010guaranteed}. Eq.\eqref{eq:dai_opt} can be solved efficiently using the ADMM optimization strategy \cite{boyd2011distributed}. \cite{dai2014simple} optimizes Eq.\eqref{eq:dai_opt} by penalizing each singular value of $\mathbf{X}^{\sharp}$ equally. Yet, we have an initial shape prior $\mathbf{X}_{init}$ that we can exploit to recover a better shape. In the same vein, Kumar \cite{kumar2020non} introduced WNN minimization to Eq.\eqref{eq:dai_opt}, which shows highly effective results with the use of $\mathbf{X}_{init}$ singular values as prior. \cite{kumar2020non} suggested the following changes
\begin{equation}\label{eq:kumar_change}
\begin{aligned}
& \displaystyle \underset{ \mathbf{X}^{\sharp},  \mathbf{X}} {{\textrm{minimize}}} ~\frac{1}{2}\|\mathbf{W} - \mathbf{R}\mathbf{X}\|_\mathcal{F}^{2} + \mu\| \mathbf{X}^{\sharp}\|_{*, \theta}; {\textrm{subject to:}}~\mathbf{X}^{\sharp} = \mathbf{\Phi}(\textbf{X})
\end{aligned}
\end{equation}
Here, $\theta$ is the weight assigned to $\mathbf{X}^{\sharp}$  based on the  $\mathbf{X}_{init}^{\sharp}$ singular values. It is known that for a low-rank shape matrix, a few top singular values contain most of the shape information. Thus, when optimizing Eq.\eqref{eq:kumar_change} the first singular value should be penalized the least and vice-versa,  using the following relation 
\begin{equation}\label{eq:weight_init}
\theta_{i} = {\xi}{\big({\sigma_{i}(\mathbf{X}_{init}^{\sharp}) + \gamma}\big)^{-1}}
\end{equation}
where, $\xi$ and $\gamma$ are small positive scalars, $ \sigma_i(\mathbf{X}_{init}^{\sharp})$ denotes the $i^{th}$ singular value of $\mathbf{X}_{init}^{\sharp}$ and $\theta_{i}$ denotes its weight. It is observed and empirically tested that such a strategy provides significantly better minima after optimization \cite{kumar2020non}.

Nevertheless, for shape estimation, contrary to Kumar\cite{kumar2020non}, we propose the mixed use of partial sum minimization of singular values and weighted nuclear norm optimization of $\mathbf{X}^{\sharp}$. Based on our extensive empirical study over several non-rigid shapes, we found that the first singular value of $\mathbf{X}_{init}^{\sharp}$ contains useful information. Thus, penalizing it during WNN minimization may hurt performance unnecessarily. Therefore, we propose to preserve the first singular value of the shape during optimization, leading to the following optimization problem

\begin{equation}\label{eq:new_opt_pssv}
\begin{aligned}
& \displaystyle \underset{ \mathbf{X}^{\sharp},  \mathbf{X}} {{\textrm{minimize}}} ~\frac{1}{2}\|\mathbf{W} - \mathbf{R}\mathbf{X}\|_\mathcal{F}^{2} + \mu\| \mathbf{X}^{\sharp}\|_{r=N, \theta}; ~{\textrm{subject to:}}~ \mathbf{X}^{\sharp} = \mathbf{\Phi}(\textbf{X})
\end{aligned}
\end{equation}
We use $N=1$ for all our experiments and assign the weights $\theta$ using Eq.\eqref{eq:weight_init} for the rest of the singular values in the shape matrix optimization via ADMM \cite{boyd2011distributed}.

\smallskip
\formattedparagraph{Shape optimization}. We optimized Eq.\eqref{eq:new_opt_pssv}  using ADMM \cite{boyd2011distributed}. Introducing the Lagrange multiplier in Eq.\eqref{eq:new_opt_pssv} gives us
\begin{equation}\label{eq:pssv_lagrange_form}
\begin{aligned}
& \displaystyle \mathcal{L}_{\rho}(\mathbf{X^{\sharp}}, \mathbf{X}) =  \frac{1}{2}\|\mathbf{W} - \mathbf{R} \mathbf{X}\|_\mathcal{F}^{2} + \mu \|\mathbf{X^{\sharp}}\|_{r=N, \theta} \\ 
& \displaystyle + \frac{\rho}{2}\|\mathbf{X}^{\sharp} - \mathbf{\Phi}(\mathbf{X})\|_\mathcal{F}^2 + <\mathbf{Y}, \mathbf{X}^{\sharp}-\mathbf{\Phi}(\mathbf{X})>
\end{aligned}
\end{equation}

\noindent
$\mathbf{Y} \in \mathbb{R}^{F \times 3P}$ is the Lagrange multiplier and $\rho > 0$ is the penalty parameter. We obtain the solution to each variable solving the following sub-problems over iterations (indexed with the variable $t$):

\begin{equation}\label{eq:pssv_update}
\begin{aligned}
& \displaystyle \mathbf{X}_{t+1} = \underset{\mathbf{X}}{\text{argmin}} ~\mathcal{L}_{\rho t}\big(\mathbf{X^{\sharp}}, \mathbf{X}_t\big); ~\mathbf{X}^{\sharp}_{t+1}= \underset{\mathbf{X^{\sharp}}}{\text{argmin}} ~\mathcal{L}_{\rho t}\big(\mathbf{X}^{\sharp}_t, \mathbf{X}\big)
\end{aligned}
\end{equation}
Using Eq.\eqref{eq:pssv_lagrange_form}-Eq.\eqref{eq:pssv_update}, we derive the following expression for $\mathbf{X}$, assuming $\mathbf{X}^{\sharp}$ is constant.
\begin{equation}\label{eq:pssv_x_solution}
\begin{aligned}
& \displaystyle \mathbf{X} \simeq 
\underset{\mathbf{X}}{\text{argmin}} \frac{1}{2}\| \mathbf {W} - \mathbf{R} \mathbf{X}\|_\mathcal{F}^{2}  + \frac{\rho}{2} \Big \| \mathbf{X} - \Big(\mathbf{\Phi}^{-1}(\mathbf{X}^{\sharp}) + \frac{\mathbf{\Phi}^{-1}(\mathbf{Y})}{\rho}\Big)\Big\|_\mathcal{F}^{2}
\end{aligned}
\end{equation}
The closed form solution for $\mathbf{X}$ is obtained by taking the derivative of Eq.\eqref{eq:pssv_x_solution} w.r.t the corresponding variable and equating it to zero. The closed form expression is used during the ADMM iteration until convergence to recover optimal $\mathbf{X}$. Similarly, rewriting the Eq.\eqref{eq:pssv_lagrange_form} by assuming $\mathbf{X}^{\sharp}$ as variable and $\mathbf{X}$ as constant, we get the following expression for $\mathbf{X}^{\sharp}$
\begin{equation}\label{eq:pssv_ssharp}
\begin{aligned}
\displaystyle \mathbf{X}^{\sharp} \simeq \underset{\mathbf{X}^{\sharp}}{{\text{argmin}}} ~\mu \|\mathbf{X}^{\sharp}\|_{r=N, \theta} + \frac{\rho}{2}\Big\|\mathbf{X^{\sharp}} - \Big(\mathbf{\Phi}(\mathbf{X}) - \frac{\mathbf{Y}}{\rho}\Big) \Big\|_\mathcal{F}^2
\end{aligned}
\end{equation}

\noindent
To solve Eq.\eqref{eq:pssv_ssharp}, we used the theory of Partial Singular Value Thresholding (PSVT) \cite{oh2016partial}. Let $\mathcal{P}_{N, \tau}[\mathbf{Q}]$ denote the PSVT operator operating on matrix $\mathbf{Q}$. The operator preserves the leading $N$ singular values and penalizes the others with soft-thresholding parameter $\tau$ \footnote{For more discussion on partial sum minimization of singular values, cf. the supplementary material. For a comprehensive theory refer to \cite{oh2016partial}.}. For completeness, let's go over the following:
\begin{table*}[t]
\centering
\resizebox{\textwidth}{!}
{
\scriptsize
\begin{tabular}{>{\columncolor[gray]{0.75}}c|c|c|c|c|c|c|c|c|c|>{\columncolor{red!15}}c}
\hline
\rowcolor{amber!50}
{\scriptsize{Dataset$\downarrow \slash$ Method$\rightarrow$}} & {\scriptsize{MP}}\cite{paladini2009factorization} & {\scriptsize{PTA}}\cite{akhter2009nonrigid}  & {\scriptsize{CSF1}}\cite{gotardo2011computing} & {\scriptsize{CSF2}}\cite{gotardo2011non} & {\scriptsize{KSTA}}\cite{gotardo2011kernel} & {\scriptsize{PND}}\cite{lee2013procrustean} & {\scriptsize{CNS}} \cite{lee2016consensus} & {\scriptsize{BMM}}\cite{dai2014simple} & {R-BMM}\cite{kumar2020non} & {\cellcolor{red!20} \scriptsize{Ours} }\\ \hline
Drink  &  0.0443 & 0.0250   & 0.0223 & 0.0223   & 0.0156 &  \textbf{0.0037}  & 0.0431 & 0.0266 & 0.0119 & \underline{\textbf{\textcolor{blue}{0.0071}}} ($K=12$)  \\ \hline
Pickup &  0.0667 & 0.2369   & 0.2301 & 0.2277   & 0.2322 &  0.0372   & 0.1281 & 0.1731  & {0.0198} & \textbf{0.0152} ($K=12$) \\ \hline
Yoga   &  0.2331 & 0.1624   & 0.1467 & 0.1464   & 0.1476 &  0.0140   & 0.1845 & 0.1150  & {0.0129} & \textbf{0.0122} ($K=10$) \\ \hline
Stretch & 0.2585 & 0.1088   & 0.0710 & 0.0685   & 0.0674 &  0.0156   & 0.0939 & 0.1034  & {0.0144} & \textbf{0.0124} ($K=11$) \\ \hline
Dance   & 0.2639 & 0.2960   & 0.2705 & 0.1983   & 0.2504 &  0.1454   & \textbf{0.0759} & 0.1864  & 0.1491  & \underline{\textbf{\textcolor{blue}{0.1209}}} ($K=4$) \\ \hline
Face   &  0.0357 & 0.0436   & 0.0363 & 0.0314   & 0.0339 &  0.0165   & 0.0248 & 0.0303  & {0.0179} & \textbf{0.0145} ($K=7$)  \\ \hline
Walking & 0.5607 & 0.3951   & 0.1863 & 0.1035   & 0.1029 &  {0.0465}   & \textbf{0.0396} & 0.1298  & 0.0882 & 0.0816 ($K=8$)  \\ \hline
Shark  & 0.1571 & 0.1804   & \textbf{0.0081} & 0.0444   & 0.0160 & {0.0135}   & 0.0832 & 0.2311  & 0.0551 & 0.0550 ($K=3$)  \\ \hline
\end{tabular}
}
\caption{Statistical comparison on the MoCap dataset \cite{akhter2009nonrigid}.  Our method provides favorable 3D reconstruction results. Contrary to the R-BMM \cite{kumar2020non}, our approach provides a methodical way to solve \nrsfm ~factorization irrespective of the camera motion assumption. The value of $K$ used is generally same as \cite{dai2012simple, dai2014simple}.  The $2^\text{nd}$ best results are underlined. \texttt{To have clear spacing, we put comparison with other methods as suggested by the reviewers in the supplementary material}.} \label{tab:cmu_statistical_results}
\end{table*}

\begin{theorem}
Oh \etal \cite{oh2016partial} proposed the following optimization problem to solve 
\begin{equation}
    \underset{\mathbf{P}}{{\emph{argmin}}} ~\tau\|\mathbf{P}\|_{r=N} + \frac{1}{2} \|\mathbf{P}-\mathbf{Q}\|_\mathcal{F}^2 
\end{equation}
where, $\tau > 0$ and $\mathbf{P}$, $\mathbf{Q}$ $\in \mathbb{R}^{m \times n}$ be real valued matrices which can be decomposed by Singular Value Decomposition (SVD). Then, the optimal solution can be expressed by the PSVT operator defined as:
\begin{equation}
    \displaystyle \mathcal{P}_{N, \tau}[\mathbf{Q}] = \mathbf{U}_{Q}(\mathbf{\Sigma}_{Q1} + \mathcal{S}_{\tau}[\mathbf{\Sigma}_{Q2}])\mathbf{V}_{Q}^{T}
\end{equation}
where, $\mathbf{\Sigma}_{Q1} = \mathbf{diag}(\sigma_1,\sigma_2,..,\sigma_N,..,0)$ and $\mathbf{\Sigma}_{Q2} = \mathbf{diag}(0,..\sigma_{N+1},..,\max(m, n))$.  Symbol $\mathcal{S}_{\tau}$ is the soft-thresholding operator defined as  $\mathcal{S}_{\tau}(\sigma) = \emph{sign}(\sigma)\emph{max}(|\sigma|-\tau, 0)$. $\mathbf{Q} = \mathbf{U}_{Q} (\mathbf{\Sigma}_{Q1} + \mathbf{\Sigma}_{Q2})\mathbf{V}_{Q}^{T}$
\end{theorem}

\noindent
For a detailed derivation and proof, we refer to Oh \etal~\cite{oh2016partial}. Using the theorem, we substitute $N=1$, $\tau = (\mu\theta)/\rho$ and write the solution of $\mathbf{X}^{\sharp}$ in Eq.\eqref{eq:pssv_ssharp} as: 
\begin{equation}
    \mathbf{X}^{\sharp} = \mathcal{P}_{1, \frac{\mu \theta}{\rho}}\Big[\big(\mathbf{\Phi}(\mathbf{X}) - {\rho}^{-1}\mathbf{Y}\big)\Big]
\end{equation}

\noindent
We use the above expression of $\mathbf{X}^{\sharp}$ during the ADMM optimization \cite{boyd2011distributed} to recover the optimal shape matrix. The $\theta$ values are assigned according to Eq.\eqref{eq:weight_init} for $N>1$. The Lagrange multiplier $(\mathbf{Y})$ and penalty parameter $(\rho)$ are updated over ADMM iteration (say for $t+1$ iteration) as $\mathbf{Y}_{t+1} = \mathbf{Y}_{t} + \rho(\mathbf{X}^{\sharp}_{t+1} - \mathbf{\Phi}(\mathbf{X}_{t+1})); ~\rho_{t+1} = {\text{minimum}}(\rho_\textrm{max}, \lambda\rho_{t})$.
Where, $\rho_\textrm{max}$ refers to the maximum value of `$\rho$' and $\lambda$ is an empirical constant. $\mathbf{Y}$ and $\rho$ are updated during the ADMM \cite{boyd2011distributed} iteration until convergence criteria is satisfied. The criteria for the ADMM iteration to stop are $\|\mathbf{X}^{\sharp} - \mathbf{\Phi}(\mathbf{X})\|_{\infty} < \epsilon, ~\text{or}, ~\rho_{t+1} \geq \rho_\textrm{max}$

\begin{figure*}[t]
\centering
\includegraphics[width=1.0\textwidth] {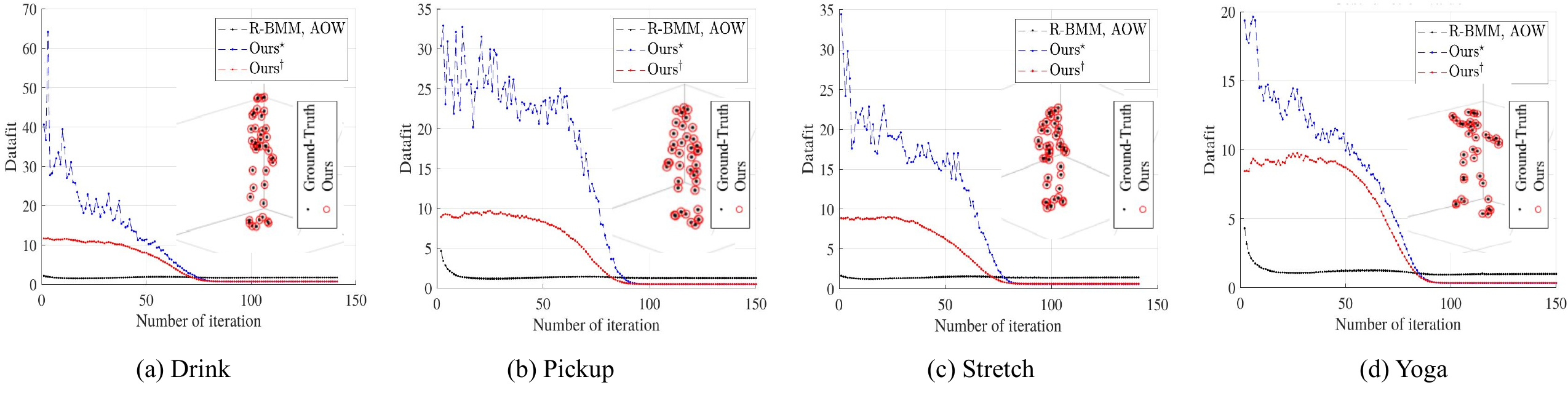}
\caption{Datafit curve shows the value of Eq.\eqref{eq:new_opt_pssv} cost function over iteration compared to R-BMM \cite{kumar2020non} and AOW \cite{iglesias2020accurate} WNN formulation on Mocap dataset \cite{akhter2009nonrigid}. Ours$^{\star}$ show the Eq.\eqref{eq:new_opt_pssv} datafit value using \cite{dai2014simple} rotation whereas, Ours$^{\dagger}$ show the datafit curve using our rotation initialization. Our shape optimization gives better minima, and using our rotation as initialization, we have faster and stable convergence response (Ours$^{\dagger}$).}
\label{fig:ours_results_cmu}
\end{figure*}

\section{Experiments}\label{sec:experiments}
\formattedparagraph{(a) Implementation Details and Initialization.} We implemented our method on a desktop machine with 32GB RAM using C++/MATLAB software. Initial rotation filtering parameter $\delta$ is set to $0.05$. All the $K$ rotations $\mathbf{R}^{k} \in \mathbb{R}^{2F \times 3}$ can be solved in parallel, so the minor increase in processing time compared to \cite{dai2014simple}, is due to registration, filtering and rotation averaging. For \eg, a 357 frame pickup sequence takes 7.46s. for registration, 0.012s. for filtering and 0.93s. for computing $\mathbf{R}_\text{avg}$. 
We ran 50 iterations of SRA (\textbf{Algorithm} \ref{alg:single_motion_avg_algo}) for each frame. The weights ($\theta$) in Eq.\eqref{eq:new_opt_pssv} is initialized using Eq.\eqref{eq:weight_init} with $\xi = 5e^{-3}*\text{sqrt}({\sigma_1(\mathbf{X}_\textit{init}^{\sharp})})$ generally and $\gamma = 1e^{-6}$. 
For Eq.\eqref{eq:new_opt_pssv} optimization via ADMM, we use $\rho = 1e^{-4}$, $\lambda = 1.1$, $\rho_\text{max} = 1e^{10}$, $\mathbf{Y} = \text{zeros}(F, 3P)$, $\epsilon_{t} = 1e^{-10} $, and $\epsilon = 1e^{-10}$ (see supplementary pseudo code). 

\smallskip
\formattedparagraph{(b) Evaluation Metric.} 
We used the popular mean normalized 3D reconstruction error metric to report our statistical results on motion capture (MoCap) benchmark \cite{akhter2009nonrigid,torresani2008nonrigid} and Garg \etal \cite{garg2013dense} dense \nrsfm ~benchmark dataset. It is defined as $e_{3d} = \frac{1}{F} \sum_{i=1}^{F} \|\mathbf{X}_{i}^{est} - \mathbf{X}_{i}^{gt} \|_\mathcal{F}/{\|\mathbf{X}_{i}^{gt}\|_\mathcal{F}}$ with $\mathbf{X}_{i}^{est}$, $\mathbf{X}_{i}^{gt}$ symbolizing per frame estimated shape and its ground-truth (GT) value, respectively.  For evaluation on recent \nrsfm ~benchmark dataset \cite{jensen2018benchmark}, we used their supplied  error evaluation metric script, which is inspired from Taylor \etal ~work \cite{taylor2010non}. The 3D reconstruction accuracy is computed after registering the recovered shape to the ground-truth shape due to global ambiguity \cite{akhter2009nonrigid,jensen2018benchmark}. To evaluate rotation estimate accuracy, we use the mean rotation error metric $e_\mathbf{R} = \frac{1}{F} \sum_{i=1}^{F} \| \mathbf{R}_{i}^\text{GT} - \mathbf{R}_{i}^\text{est} \|_\mathcal{F}$. Here, $\mathbf{R}_{i}^\text{GT}$, $\mathbf{R}_{i}^\text{est}$ denotes the ground-truth and estimated per frame rotation.

\subsection{Dataset and Evaluation.}
\formattedparagraph{(a) MoCap Benchmark Dataset.}  Introduced by Akther \etal \cite{akhter2009nonrigid} and Torresani \etal \cite{torresani2008nonrigid}, this dataset has become a standard benchmark for any \nrsfm ~algorithm evaluation. It is composed of 8 real sequences, namely Drink (1102, 41), Pickup (357, 41), Yoga (307, 41), Stretch (370, 41), Dance (264, 75), Walking (260, 55), Face (316, 40) and Shark (240, 91). The last 3 sequences were introduced by Torresani \etal \cite{torresani2008nonrigid}. The numbers presented in bracket correspond to number of frames and points $(F, P)$. Tab.(\ref{tab:cmu_statistical_results}) shows the comparison of our method with other competing methods. For evaluation, we keep the value of $K$ generally same as BMM \cite{dai2014simple}. From Tab.(\ref{tab:cmu_statistical_results}), it is easy to observe that more often than not, our approach performs best or second-best than other methods, thus showing a consistent superior performance over a diverse set of object deformation type. 

Compared to BMM \cite{dai2014simple}, which also makes no assumption other than low-rank, our \textcolor{red}{{{\textless\textless}}}organic prior\textcolor{red}{{{\textgreater\textgreater}}} based method dramatically improves 3D reconstruction accuracy, thereby validating our claims made in the paper. Fig.(\ref{fig:ours_results_cmu}) shows few qualitative results along with the convergence curve comparison with the current methods such as R-BMM \cite{kumar2020non}, AOW \cite{iglesias2020accurate}. The results show recovery of better minima and stable convergence curve using our rotation estimate initialization.

\begin{table*}[t]
\footnotesize
\centering
\resizebox{\textwidth}{!}
{
\begin{tabular}{c|c|c|c|c|c|>{\columncolor{red!15}}c|c|c|c|c|c|>{\columncolor{red!15}}c}
    \hline
    \rowcolor[gray]{0.80}
    \multicolumn{1}{c|}{Method Type $\rightarrow$} & \multicolumn{6}{c|}{\cellcolor{amber!50} \textbf{Sparse \nrsfm ~Methods} } & \multicolumn{6}{c}{\cellcolor{amber!50} \textbf{Dense \nrsfm ~Methods} }  \\
    \hline
        Dataset &  MP \cite{paladini2009factorization} &  PTA \cite{akhter2009nonrigid}  & CSF1 \cite{gotardo2011computing} &  CSF2 \cite{gotardo2011non}  & BMM \cite{dai2014simple} & Ours & DV \cite{garg2013dense} &  SMSR \cite{ansari2017scalable} &  CMDR \cite{golyanik2020intrinsic} &  GM \cite{kumar2018scalable} &  ND \cite{sidhu2020neural} & Ours \\
        \hline
        Face Seq.1  &  0.0926 &  0.1559 & 0.5325  & 0.4677 & 0.4263  & \textbf{0.0624} & 0.0531 & 0.1893 & - & \textbf{0.0443} & - & 0.0624   \\
        \hline
        Face Seq.2  & 0.0819 & 0.1503 & 0.9266  & 0.7909 & 0.6062  & \textbf{0.0451} &  0.0457  & 0.2133 & - & \textbf{0.0381} & - & \underline{\textbf{\textcolor{blue}{0.0451}}} \\
        \hline
        Face Seq.3  & 0.1057 & 0.1252 & 0.5274  &  0.5474 & 0.0784   & \textbf{0.0279} & 0.0346 & 0.1345 & 0.0373 & 0.0294 & 0.0450 & \textbf{0.0279} \\
        \hline
        Face Seq.4  & 0.0717 &  0.1348 &  0.5392  & 0.5292 & 0.0918  & \textbf{0.0419} & 0.0379 & 0.0984 & 0.0369 & \textbf{0.0309} & 0.0490 & 0.0419\\
    \hline
\end{tabular}
}
\caption{3D reconstruction accuracy on dense \nrsfm ~dataset \cite{garg2013dense}.  We observed superior results compared to the well-known sparse \nrsfm ~methods.  It is interesting to observe that our results compares favorably to carefully crafted dense \nrsfm ~methods such as DV, GM and others. The $2^\textrm{nd}$ best performance of our method is underlined. }
\label{tab:dense_nrsfm_results}
\end{table*}

\smallskip
\formattedparagraph{(b) Dense \nrsfm ~Benchmark Dataset.} Introduced by Garg \etal \cite{garg2013dense,garg2013variational}, it is a standard dataset to evaluate dense \nrsfm ~methods. It comprises of 4 synthetic face sequences and 3 real video sequences of heart, back, and face deformation. The synthetic face dataset is composed of 28,880 tracked feature points. Face sequence 1 and Face sequence 2 are 10 frames long video, whereas Face sequence 3 and Face sequence 4 are 99 frames video. The video sequence for heart, back, and face dataset is 80, 150, and 120 frames long with 68295, 20561, and 28332 feature track points. Tab.(\ref{tab:dense_nrsfm_results}) provides the statistical results of our approach compared to well-known dense \nrsfm ~algorithms. For better comprehension, we classified the comparison into two sets \ie, sparse \nrsfm ~methods and dense \nrsfm ~methods. From Tab.(\ref{tab:dense_nrsfm_results}), it is easy to observe the advantage of our approach compared to well-known sparse \nrsfm ~methods. For evaluation of our method, we use $K=1$ for all the four sequence. For other methods \cite{dai2014simple,akhter2009nonrigid}, we iterate over different $K$ and put its best possible results.

The interesting point to note is that without using any extra assumptions about the dense deforming surface such as union of linear subspaces \cite{kumar2018scalable, kumar2019jumping}, variation in the deformation over frame should be smooth \cite{garg2013variational}, dynamic shape prior \cite{golyanik2020intrinsic}, smooth trajectory constraint \cite{ansari2017scalable}, and recent deep neural network based latent space constraint \cite{sidhu2020neural}, our method provide impressive results and it is close to the best method\cite{kumar2018scalable}.  Note that, contrary to our simple approach, GM\cite{kumar2018scalable} is a complex geometric method to implement. To conclude, our results reveal the strength of classical \nrsfm ~factorization if organic priors are exploited sensibly.

\begin{figure*}[t]
\centering
\includegraphics[width=1.0\textwidth] {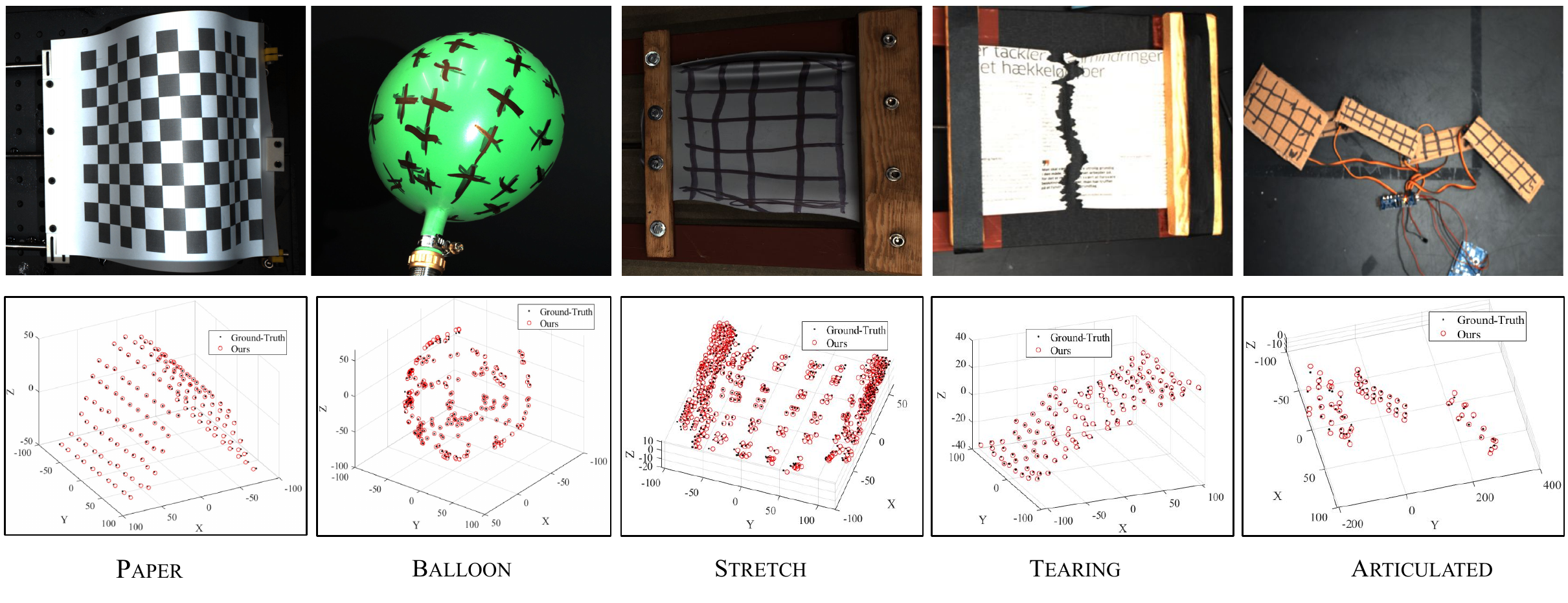}
\caption{Qualitative results on \nrsfm ~challenge dataset \cite{jensen2018benchmark}. \textbf{Top row.} Subject image. \textbf{Bottom row.} 3D reconstruction of the respective object shape. }
\label{fig:nrsfm_challenge_visual}
\end{figure*}

\smallskip
\formattedparagraph{(c) \nrsfm ~Challenge Dataset.} 

\setlength{\intextsep}{1pt}%
\setlength{\columnsep}{8pt}%
\begin{wraptable}{r}{8.5cm}
\scriptsize
\centering
\begin{tabular}{c|c|c|c|c|>{\columncolor{red!15}}c}
\hline
\rowcolor[gray]{0.75}
Data & BMM \cite{dai2014simple} & R-BMM \cite{kumar2020non} & AOW \cite{iglesias2020accurate} & BP \cite{ornhag2021bilinear} & Ours \\ \hline
Articul. &  18.49   & 16.00   & 15.03   & 16.10 & \textbf{12.18} $(K=8)$ \\ \hline
Balloon &  10.39   & 7.84    &  8.05   & 8.29 & \textbf{6.29} $(K=5)$ \\ \hline
Paper &  8.94  & 10.69   & 10.45   & \textbf{6.70} & \underline{\textbf{\textcolor{blue}{8.86}}} $(K=2)$ \\ \hline
Stretch & 10.02  & 7.53   &  9.01   & 7.66 & \textbf{6.36} $(K=6)$ \\ \hline
Tearing & 14.23   & 16.34   &  16.20  & 11.26 & \textbf{10.91} $(K=6)$ \\ \hline
\end{tabular}
\caption{\footnotesize Comparison of our method with state-of-the-art on recent benchmark \cite{jensen2018benchmark}. Results are reported in millimeters.}\label{tab:nrsfm_challenge}
\end{wraptable}

\noindent
Jensen \etal \cite{jensen2018benchmark} recently proposed this dataset. It comprises 5 different subjects, namely Articulated, Paper, Balloon, Stretch, and Tearing. Each subject's deformations is captured under 6 varying camera trajectories \ie, circle, flyby, line, semi-circle, tricky and zigzag, making the dataset interesting yet challenging.  For evaluation, the dataset provide a single frame ground-truth 3D shape for each subject. Tab.(\ref{tab:nrsfm_challenge}) show the average 3D reconstruction accuracy comparison in millimeters with the recent and earlier state-of-the-art on this dataset \ie, BMM \cite{dai2014simple}, R-BMM\cite{kumar2020non}, AOW \cite{iglesias2020accurate}, BP\cite{ornhag2021bilinear}.  For comparison, we used the orthogonal sequence of the dataset. The value of $K$ used by our method for comparison is provided in the bracket. Statistical results indicate that our approach provides better non-rigid shape reconstruction for most of the subject categories on this dataset. Fig.(\ref{fig:nrsfm_challenge_visual}) show visual results obtained on this dataset.

\smallskip
\formattedparagraph{(d) Rotation Estimation.}
To validate that the single rotation averaging gives meaningful rotation, we validate our results using the ground-truth rotation available in the Akther \etal \cite{akhter2009nonrigid} dataset. 

\begin{figure}[t]
\centering
\subfigure [\label{fig:pickup_noise_motion} $e_\mathbf{R}$ w.r.t noise]
{\includegraphics[width=0.23\textwidth, height=0.12\textheight]{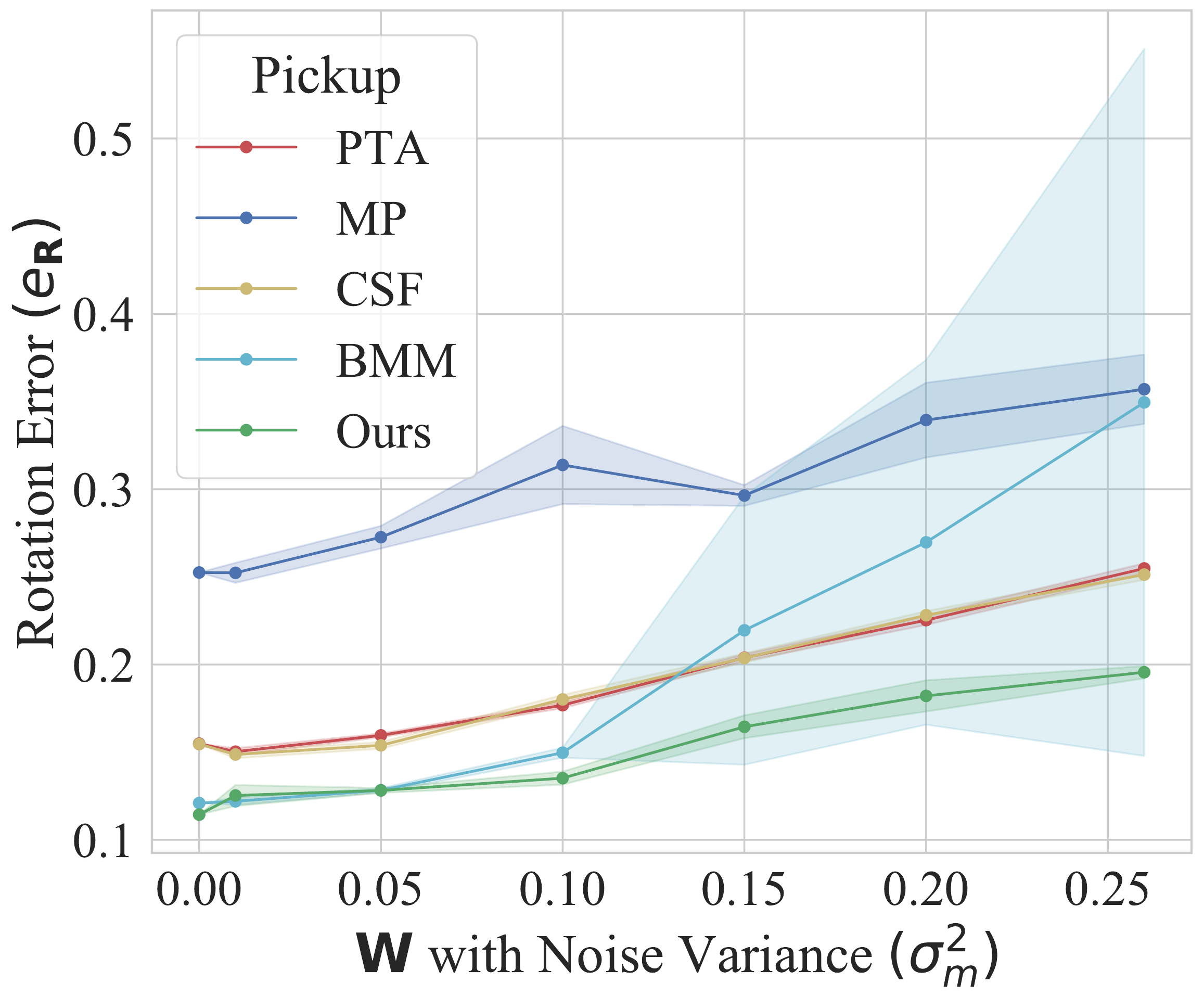}}
\subfigure [\label{fig:pickup_noise_shape} $e_{3d}$ w.r.t noise ] {\includegraphics[width=0.23\textwidth, height=0.12\textheight]{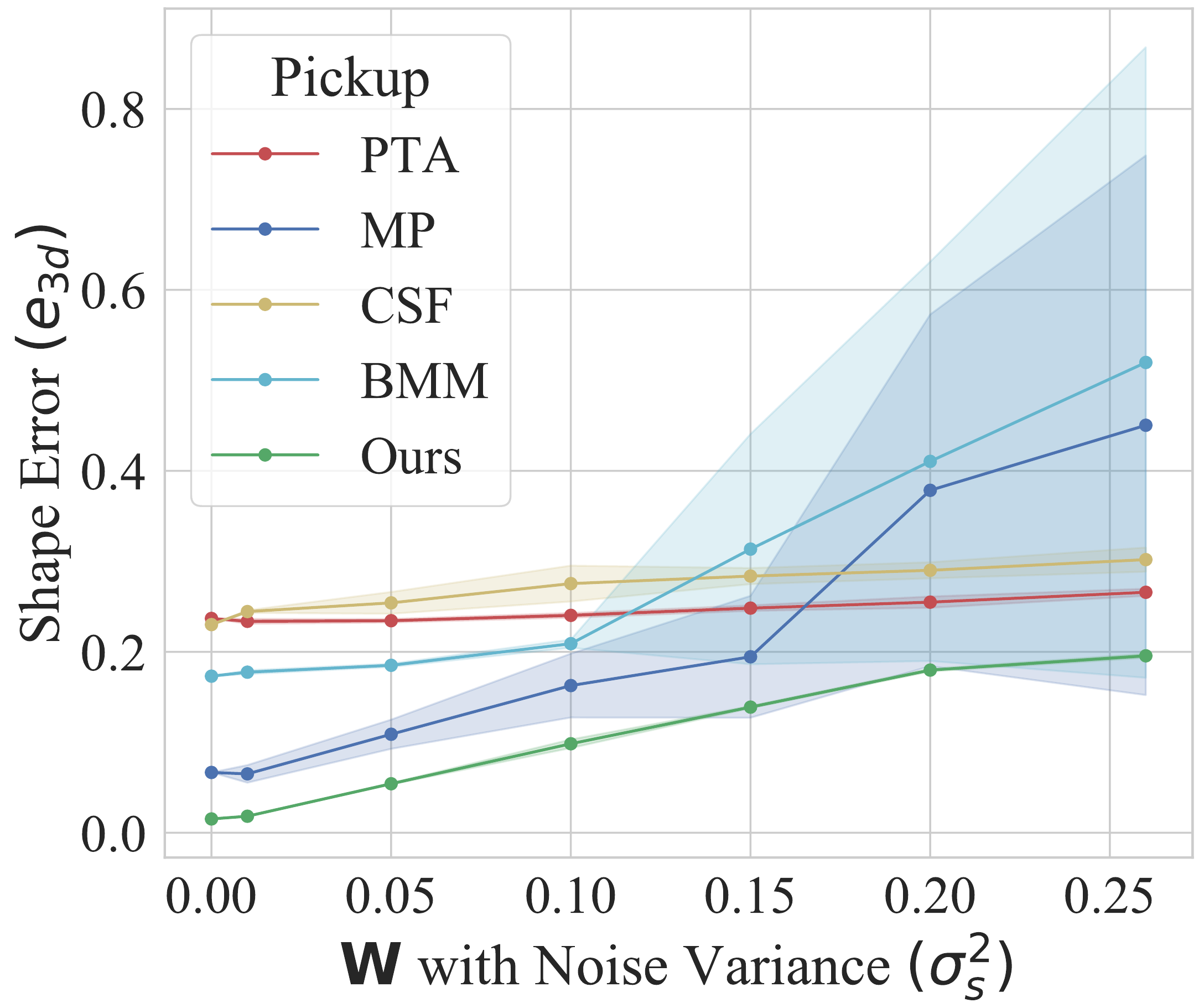}}
\subfigure [\label{fig:missing_data} missing data.] {\includegraphics[width=0.22\textwidth, height=0.12\textheight]{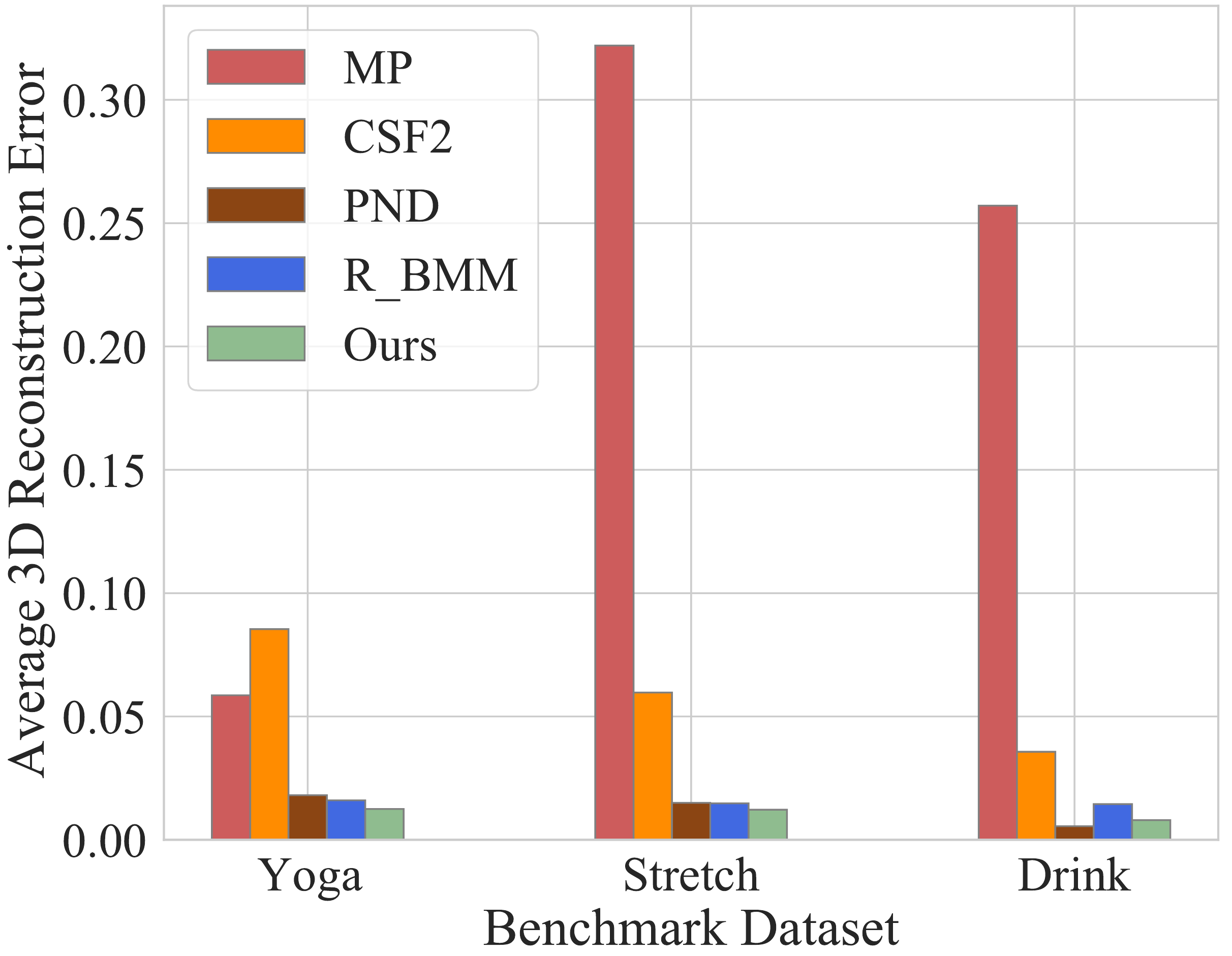}}
\subfigure [\label{fig:wrtN} $e_{3d}$ w.r.t $N$] 
{\includegraphics[width=0.25\textwidth, height=0.12\textheight]{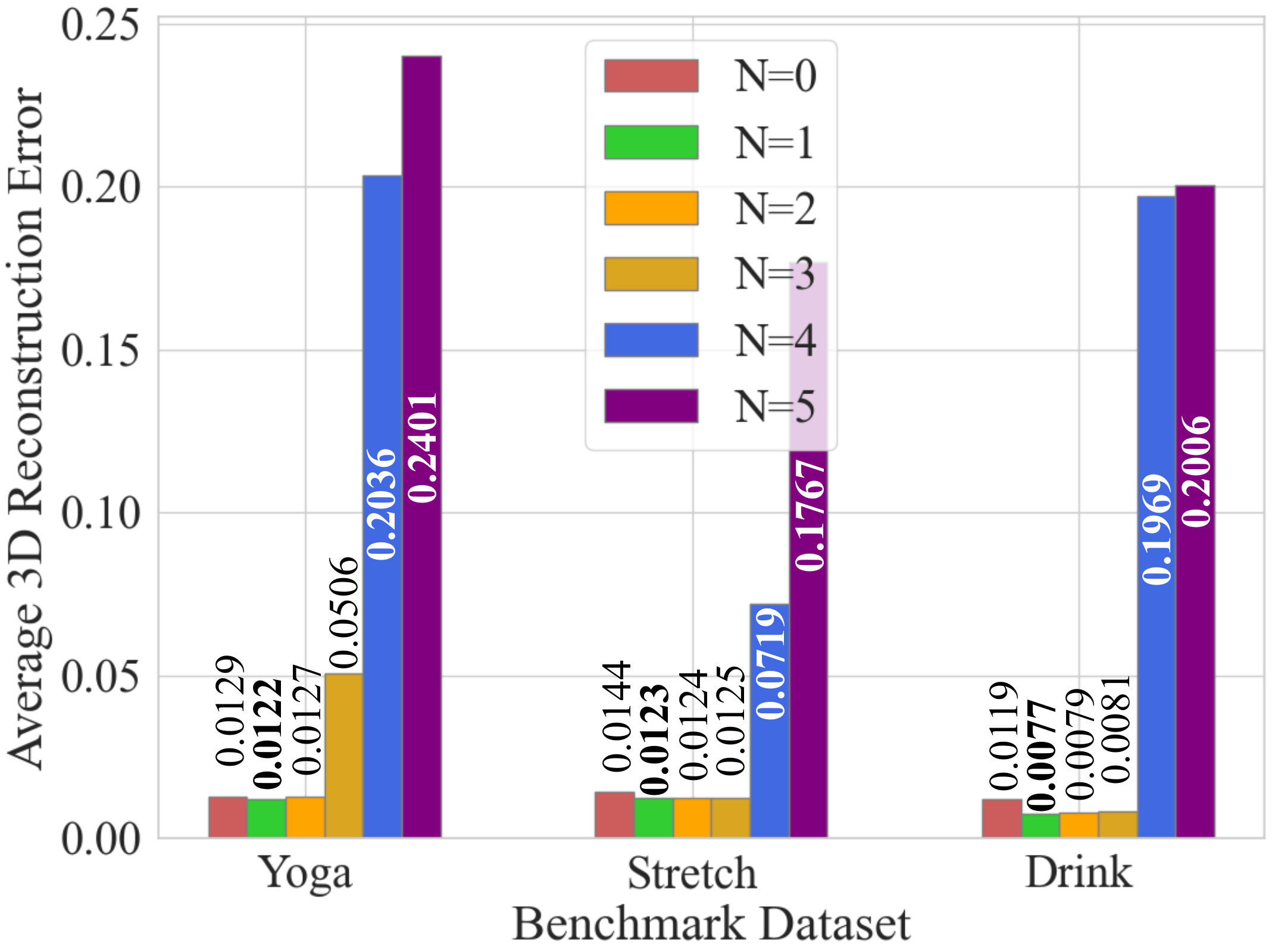}}
\caption{(a)-(b) Rotation and shape error on the noisy Pickup trajectories. Our method show stable behaviour consistently. Mean and standard deviation is shown with bold and light shaded regions, respectively. (c) Avg. 3D reconstruction for missing data experiment. (d) $N=0$ show results when all the singular values are penalized using WNN. Our approach \ie, $N=1$ gives better results overall under same rotation.}
\label{fig:evaluationNandMD}
\end{figure}

\setlength{\intextsep}{1pt}%
\setlength{\columnsep}{7pt}%
\begin{wraptable}{r}{8.5cm}
\scriptsize
\centering
\begin{tabular}{c|c|c|c|c|c|>{\columncolor{red!15}}c}
\hline
\rowcolor[gray]{0.75}
Data & MP \cite{paladini2009factorization} & PTA \cite{akhter2009nonrigid} & CSF \cite{gotardo2011computing} & BMM \cite{dai2014simple} & R-BMM \cite{kumar2020non} & Ours \\ \hline
Yoga &  0.8343   & 0.1059   & 0.1019   & 0.0883 & 0.0883 & {0.0888}  \\ \hline
Pickup &  0.2525   & 0.1549    &  0.1546   & 0.1210 & 0.1217 & \textbf{0.1144}  \\ \hline
Stretch &  0.8185  & 0.0549  & 0.0489   & 0.0676 & 0.0676 & 0.0671 \\ \hline
Drink & 0.2699  & 0.0058   &  0.0055   & 0.0071 & 0.0243 & 0.0072 \\ \hline
\end{tabular}
\caption{ \footnotesize  $e_\mathbf{R}$ comparison with other factorization methods.}\label{tab:rotation_comparison}
\end{wraptable}
Tab.(\ref{tab:rotation_comparison}) provide the average camera rotation error $e_\mathbf{R}$ results on yoga, pickup, stretch, and drink sequence. The statistics show that using our approach, we can have fine rotation estimate\footnote{With $\mathbf{W=RS}$ theory, even GT rotation cannot provide GT shape, cf. \cite{dai2014simple} Table(3)}. Further, advantage of our rotation estimation on clean sequence, noisy trajectories and pseudo inverse solution can be inferred from Tab.(\ref{tab:rotation_ablation}) Fig.\ref{fig:pickup_noise_motion}, and Tab.(\ref{tab:pinv_soln}), respectively.


\begin{table}[h]
    \scriptsize
    \centering
    \resizebox{\columnwidth}{!}
    {
    \begin{tabular}{c|c|c|c|c|c|c|c|c}
    \hline
    \rowcolor[gray]{0.75}
         Dataset & Drink & Pickup & Yoga & Stretch & Dance & Face & Walking & Shark\\
         \hline
         BMM \cite{dai2014simple} & 0.0266 & 0.1731 & 0.1150 & 0.1034 & 0.1864 & 0.0303 & 0.1298 & 0.2357\\
         \hline 
         $e_{3d}$ (\cite{dai2014simple} rotation)  & 0.0101 & 0.0164 & 0.0126 & 0.0126 & 0.1382 & 0.0152 & 0.0880 & 0.0563 \\
         \hline 
         \rowcolor{red!15}
         $e_{3d}$  (our rotation) & \textbf{0.0071} & \textbf{0.0152} & \textbf{0.0122} & \textbf{0.0124} & \textbf{0.1209} & \textbf{0.0145}  & \textbf{0.0816} & \textbf{0.0550} \\
         \hline 
    \end{tabular}
    }
    \caption{\textbf{$2^\text{nd}$ row}: Our $e_{3d}$ results using Dai \etal ~rotation\cite{dai2014simple}. \textbf{$3^\text{rd}$ row}: $e_{3d}$ using our rotation. Indeed using organic rotation priors help improve overall performance.}
    \label{tab:rotation_ablation}
\end{table}

\smallskip
\formattedparagraph{(e) Other Experiments and Ablations}.

\smallskip
\noindent
\textit{(\textbf{i}) {Performance with noisy trajectory.}} Fig.\ref{fig:pickup_noise_motion}, Fig.\ref{fig:pickup_noise_shape} shows the rotation ($e_\mathbf{R}$) and shape error ($e_{3d}$) comparison on the noisy trajectory, respectively. We introduce noise to the 2D point trajectory with the standard deviation varying from 0.01-0.25 using \texttt{normrand()} function from MATLAB. We ran the different methods 10 times for each standard deviation value and plotted the method's mean and variance. Statistical results show that our method is quite robust to noisy sequence and show much stable behaviour (both in rotation and shape estimation) than the other prior or prior-free approaches (see Fig.\ref{fig:pickup_noise_motion}-\ref{fig:pickup_noise_shape}).

\smallskip
\noindent
\textit{(\textbf{ii}) {Performance on missing trajectory cases.}}
For this experiment, we used Lee \etal \cite{lee2013procrustean} and Kumar \cite{kumar2020non} setup, where we randomly set 30\% of the trajectory missing from the $\mathbf{W}$. We perform matrix completion using \cite{cabral2013unifying} optimization and then ran our algorithm on the recovered matrix. The results are shown in Fig.\ref{fig:missing_data}. Our method outperforms the state-of-the-art in most of the cases.

\smallskip
\noindent
\textit{(\textbf{iii}) {Performance with change in value of $N$.}} To show that $N=1$ generally works best for Eq.\eqref{eq:new_opt_pssv}, we conducted this experiment. First, we penalize all the singular values using WNN optimization ($N=0$) and then we vary the value of $N$ from 1 to $5$ and recorded the results. Fig.\ref{fig:wrtN} shows the reconstruction results using different values of $N$. We observed that by penalizing all the singular values using WNN, we are unnecessarily hurting the performance. On the contrary, if we increase $N$ value greater than 1, {\texttt{more often than not}}, it starts to reduce the performance. {Refer supplementary material for more results and discussions}.
 
\section{Conclusion}
This work reveals organic priors for \nrsfm ~factorization irrespective of camera motion and shape deformation type. It exhibited that mindful use of such fundamental priors gives better accuracy than the prior-free methods. That said, our method uses an orthographic camera model with a low-rank shape assumption in \nrsfm. Hence, by construction, it has some limitations \emph{for} \eg, our method may perform inadequately on high perspective distortion images having large object deformation. A recent idea by Gra\ss hof et al. \cite{grasshof2022tensor} can be used to overcome such a limitation. Finally, we conclude that the clever use of organic priors with matrix factorization theory is sufficient to provide excellent 3D reconstruction accuracy for both sparse and dense \nrsfm.

\smallskip
\formattedparagraph{{Acknowledgement.}}{{~The authors thank Google for their generous gift (ETH Z\"urich Foundation, 2020-HS-411).}
}


\bibliographystyle{splncs04}
\bibliography{egbib}


\end{document}